\documentclass[letterpaper,10pt,conference]{ieeeconf}
\IEEEoverridecommandlockouts
\usepackage{amsmath,amssymb,amsfonts}
\usepackage{algorithmic}
\usepackage[pdftex]{graphicx}
\usepackage{textcomp}
\usepackage{xcolor}

\usepackage[bookmarks=true]{hyperref}

\let\labelindent\relax
\usepackage{enumitem}

\usepackage{float}
\usepackage{listings}
\usepackage{multirow}
\usepackage{xparse}

\makeatletter
\let\NAT@parse\undefined
\makeatother
\usepackage[numbers,sort,compress]{natbib}

\usepackage[all=normal,bibbreaks=tight,paragraphs=tight,floats=tight]{savetrees}

\def\arxiv{1}

\hypersetup{
colorlinks=true,
urlcolor=blue,
}

\newtheorem{theorem}{Theorem}[section]
\newtheorem{lemma}[theorem]{Lemma}

\ifx\arxiv\undefined
\newcommand{\appx}{Sec.~}
\else
\newcommand{\appx}{Appx.~}
\fi

\newcommand{\obj}[1]{#1}
\newcommand{\pred}[1]{\texttt{#1}}
\NewDocumentCommand \prop {>{\SplitList{ }} m} {\proposition #1}
\NewDocumentCommand \proposition {g g g g} {\texttt{#1}(#2
  \IfValueTF{#3}{,\,#3}{}
  \IfValueTF{#4}{,\,#4}{}
  )
}

\NewDocumentCommand \action {>{\SplitList{ }} m} {\actioncall #1}
\NewDocumentCommand \actioncall {g g g g} {#1(#2
  \IfValueTF{#3}{,#3}{}
  \IfValueTF{#4}{,#4}{}
  \texttt{)}
}

\lstdefinelanguage{pddl}{
    morekeywords={forall},
    otherkeywords={:goal},
    sensitive=true, 
    morecomment=[l]{;},
    morestring=[b]" 
}

\usepackage[font=footnotesize,justification=justified,singlelinecheck=false]{caption}
\usepackage{etoolbox}
\makeatother
\def\tablename{Table}

\title{\LARGE \bf
Grounding Predicates through Actions}

\author{Toki Migimatsu and Jeannette Bohg%
\thanks{The authors are with the Department of Computer Science, Stanford University, Stanford, CA 94309 USA (e-mail: \{takatoki,bohg\}@cs.stanford.edu).}%
}

\begin{document}
\maketitle




\begin{abstract}
Symbols representing abstract states such as ``dish in dishwasher" or ``cup on table" allow robots to reason over long horizons by hiding details unnecessary for high-level planning. Current methods for learning to identify symbolic states in visual data require large amounts of labeled training data, but manually annotating such datasets is prohibitively expensive due to the combinatorial number of predicates in images. We propose a novel method for automatically labeling symbolic states in large-scale video activity datasets by exploiting known pre- and post-conditions of actions. This automatic labeling scheme only requires weak supervision in the form of an action label that describes which action is demonstrated in each video. We use our framework to train predicate classifiers to identify symbolic relationships between objects when prompted with object bounding boxes, and demonstrate that such predicate classifiers can match the performance of those trained with full supervision at a fraction of the labeling cost. We also apply our framework to an existing large-scale human activity dataset, and demonstrate the ability of these predicate classifiers trained on human data to enable closed-loop task planning in the real world.
\end{abstract}


\section{Introduction}
\label{sec:intro}
Enabling robots to perform long horizon tasks such as preparing meals or assembling furniture is a widely studied problem. Long horizon planning is rooted in early AI work that studied how to give robots the ability to reason through symbols \cite{fikes1971strips}. Symbols allow robots to abstract away low-level details of the environment and perform logical reasoning at a higher level  \cite{konidaris2018skills,mao2019neuro,xu2019regression,garrett2021integrated}. However, giving robots the ability to perceive symbols in real-world environments is still an unsolved problem. Without some form of sensory grounding, propositions such as ``drawer is open" are simply a set of symbols that lack any actionable meaning for the robot. 
Thus, robots often execute symbolic plans without closed-loop visual feedback---if a robot fails to open a drawer, it has no way of knowing, because it does not know what ``drawer is open" looks like.
State-of-the-art methods for learning visual groundings of symbols require large amounts of annotated data \cite{lu2016visual,newell2017pixels,yang2017support,mao2019neuro}. However, obtaining annotations of symbolic states is prohibitively expensive due to the sheer number of propositions in a single image. Furthermore, densely-labeled datasets cannot easily be transferred between domains, since different planning problems often require different symbols.

\begin{figure}
    \centering
    \includegraphics[width=\columnwidth]{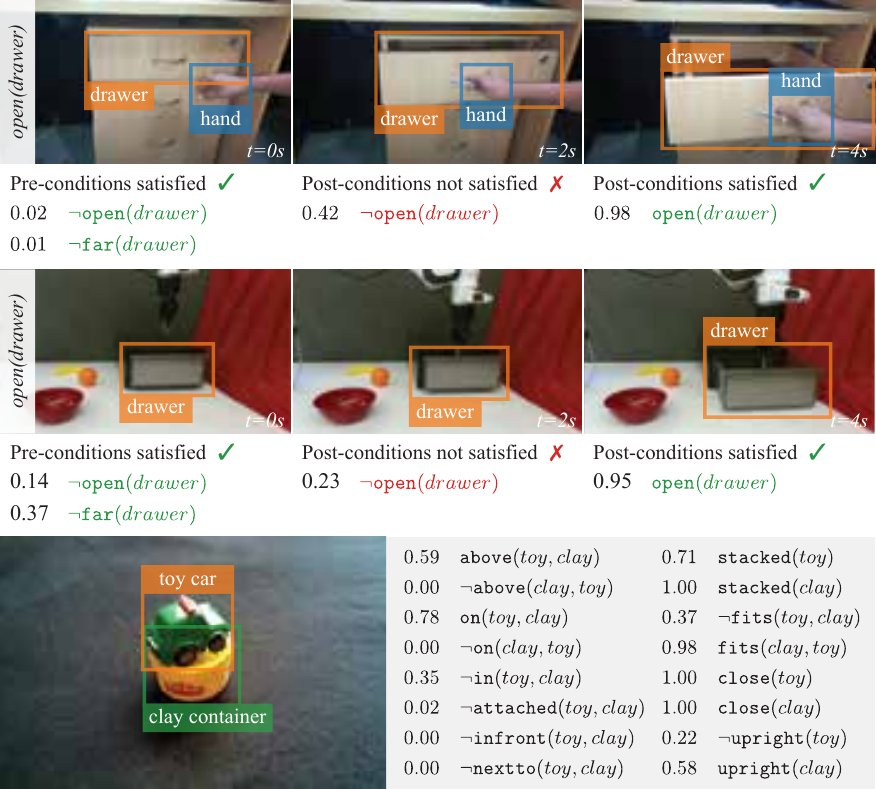}
    \caption{Example predictions of a predicate classifier trained on data labeled with our proposed method. The top two rows show how the predicate classifier can be used to determine whether the pre- or post-conditions of an action are satisfied for closed-loop task planning---the first row shows predictions on a 20BN video, and the second row shows predictions in our real robot domain. Each entry shows the predicate probability predicted by the classifier and the resulting binary classification. The red color indicates that the desired pre/post-conditions have not yet been satisfied, while green indicates that they have. The bottom shows a selection of the $151$ propositions output by the predicate classifier for one image. More examples can be found on our project website: \url{https://sites.google.com/stanford.edu/groundingpredicates}.}
    \label{fig:1-teaser}
\end{figure}

Rather than learning visual groundings from direct labels of symbolic state, we propose to learn them indirectly from visual examples of symbolic actions. Actions change the symbolic state in a predefined manner according to their pre- and post-conditions (e.g. the action ``pick up cup from table" changes the propositions ``hand is empty" and ``cup is on table" to ``hand is holding cup" and ``cup is not on table"). Labeling a dataset with actions is easier than with symbolic states: it requires first defining the pre- and post-conditions once for each action class, and then annotating each visual example with only its action. Then, partial labels of symbolic state come for free with the action pre- and post-conditions. For task planning applications, the action pre- and post-conditions will already be defined in the task planning domain.

With this novel partial labeling scheme, we train networks to infer symbolic states in images in 20BN Something Something v2 (20BN) \citep{goyal2017something}, a large-scale human activity dataset. The result is a system that can identify symbolic states in real-world environments for robot manipulation (Fig.~\ref{fig:1-teaser}). These classifiers open up many opportunities for long horizon planning in the real world, such as closed-loop task planning or learning from demonstrations of sequential actions.

The main contributions of this paper are three-fold. 1) We provide a framework for automatically labeling symbolic states in real-world image frames through actions, using the logical formalism of Planning Domain Description Language (PDDL)~\cite{mcdermott1998pddl}. 2) We evaluate this framework on two domains, 20BN and Gridworld, and analyze its advantages and disadvantages. 3) We demonstrate the ability of predicate classifiers that are only trained on human demonstrations in 20BN to be used for closed-loop task planning in a real-world robot environment.

\section{Related Work}
\label{sec:related}
\subsection{Visual Relationship Detection}

The computer vision community has a recently growing body of work on symbolic state detection, under the names \textit{visual relationship detection} and \textit{scene graph generation} \cite{lu2016visual,newell2017pixels,xu2017scene,yang2017support,liang2017deep,zhang2017visual,yang2018graph,kolesnikov2019detecting,mao2019neuro}. A scene graph represents image scenes as graphs where nodes are objects or attributes like $grass$ and $green$, and edges are relationships between nodes, e.g., $\langle grass-green \rangle$
\cite{johnson2015image}. The aim of visual relationship detection is to enable semantic scene understanding and to connect visual concepts with natural language. The three most common datasets for visual relationship detection are CLEVR (100k synthetic images with 3 objects and 16 predicates) \cite{johnson2015image}, VRD (5k real world images with 100 objects and 70 predicates) \cite{lu2016visual}, and Visual Genome (100k real world images with 34k objects and 110k predicates) \cite{krishna2017visual}. A key challenge of visual relationship detection (i.e., symbolic state detection) is the immense number of possible states. Even for the relatively small VRD dataset, 70 binary predicates with 100 objects results in $70 * 100 * 99 = 693k$ possible propositions.
VRD and Visual Genome both rely on manual annotation, and thus obtaining fully labeled symbolic states is infeasible. Our method for labeling symbolic states requires defining the pre- and post-conditions once for each action class, and then partial symbolic state labels are automatically generated for all examples in the dataset.

It is important to note that our method only works when before and after images of the actions being performed are provided; VRD and Visual Genome do not meet this requirement. Furthermore, learning from pre- and post-conditions of actions only works in domains where the symbolic state can be manipulated by actions. Propositions such as ``sky is blue" may be difficult to learn through actions
, for example. The Action Genome dataset \cite{ji2020action} introduces spatio-temporal scene graphs with the purpose of using symbolic state changes across actions to improve action recognition in videos. However, Action Genome, like VRD and Visual Genome, relies on manual annotations, and furthermore does not include negative propositions. Action Genome is based on the Charades dataset for video activity recognition \cite{sigurdsson2016hollywood}. Charades is one of numerous activity recognition datasets, like ActivityNet \cite{caba2015activitynet}, Epic-Kitchens \cite{damen2018scaling}, and Atomic Visual Actions (AVA) \cite{gu2018ava}, where labeling symbolic states with our method could be possible. However, we choose to evaluate our method on the 20BN Something Something v2 dataset \cite{goyal2017something}, a video activity recognition dataset that focuses on the manipulation of everyday objects and has been shown to be useful for learning robotic manipulation skills \cite{shao2020concept,rothfuss2018deep}.

The focus of this work is providing an alternative framework for labeling symbolic states, not proposing a novel model architecture for visual relationship detection. To evaluate our labeling method, we therefore train a model whose architecture is based on a state-of-the-art model for visual relationship detection \cite{inayoshi2020bounding}. However, our approach is agnostic to the specific model architecture.

\subsection{Visual Grounding for Planning}

A long-term goal for this work is to help bridge the gap between perception and long horizon planning in robotics. A majority of task planning methods operate solely at the level of symbolic abstractions, assuming that in a real world application, there would be some way to perceive symbols in the environment. Some works attempt to learn the symbols themselves, so that perception and planning can work together in an end-to-end fashion \cite{asai2018classical,konidaris2018skills}. In these works, the learned symbols are not easily interpretable, and guaranteeing correct behavior over a long horizon for complex domains is therefore difficult. Our labeling method benefits frameworks that rely on direct supervision of symbols to learn to plan over long horizons, such as \cite{zhu2017visual,huang2019continuous,xu2019regression,kase2020transferable,nguyen2020self}. These systems contain submodules that require annotations of predicates in images. This dependency restricts these systems to simulated environments where symbolic states are easily obtainable or to real-world domains small enough that manually annotating symbolic states is feasible.

The work most directly related to ours is perhaps an imitation learning system that learns to ground predicates associated with the pre- and post-conditions of demonstrated actions \cite{ahmadzadeh2015learning}. With grounded predicates, their system can then generalize to new tasks using classical planning. Specifically, their Visuospatial Skill Learning module learns to classify the predicate $\prop{far a}$ for a cuboid block $\obj{a}$ using demonstrations from two actions: $\action{push a}: \neg \prop{far a} \rightarrow \prop{far a}$ and $\action{pull a}: \prop{far a} \rightarrow \neg \prop{far a}$. While they learn to ground predicates from actions, the complexity of their problem is much simpler: one predicate. Our method enables learning to ground any number of predicates simultaneously from actions with arbitrarily complex pre- and post-conditions.

\subsection{Weak Supervision}

Our framework might be considered a form of weak supervision, which deals with classification problems where humans define annotation functions (e.g. pre- and post-conditions of actions) that provide noisy labels for unlabeled datasets to alleviate the effort of full labeling. Although our dataset is not strictly unlabeled (we assume action labels are available), methods from this relatively new field could be applied to our framework in future work \cite{ratner2017snorkel,varma2018snuba}. In \appx\ref{appx:calculations}, we provide back-of-the-envelope calculations to demonstrate how much annotation time weak supervision can save. Our weak supervision framework makes it easier to train predicate classifiers for custom task planning domains, which often contain symbols not transferable from other domains.

\section{Planning Domain Description Language}
\label{sec:pddl}
This paper uses the Planning Domain Description Language (PDDL) \cite{mcdermott1998pddl} to describe symbolic domains. A PDDL domain can be specified with a tuple $(\Phi, \mathcal{A})$, where $\Phi$ is the set of predicates and $\mathcal{A}$ is the set of actions. A PDDL problem is a tuple $(\mathcal{O}, s_{init}, \mathfrak{g})$, where $\mathcal{O}$ is the set of environment objects, $s_{init}$ is the initial state, and $\mathfrak{g}$ is the goal---specified as a first-order logic formula---to be satisfied.

\subsection{Predicates and Propositions}
A predicate has a fixed number of parameters, each of which can be instantiated with an object in $\mathcal{O}$ to form a proposition. For example, the predicate $\prop{in a b}$ has two parameters $a$ and $b$, and instantiating the parameters with arguments $box$ and $hand$ results in the proposition $\prop{in box hand}$. In this paper, we use ``positive proposition" to refer to atomic formulas (e.g., $\prop{in box hand}$) and ``negative proposition" to refer to negations of atomic formulas (e.g., $\neg \prop{in box hand}$). Let $\mathcal{P}$ be the set of all possible positive propositions and $N$ be the number of these propositions ($N = |\mathcal{P}|$).
\begin{align}
    \mathcal{P} = \left\{\phi(o_1, \dots, o_M) \mid \forall \phi \in \Phi, \forall (o_1, \dots, o_M) \in \mathcal{O}^M \right\}
    \label{eq:propositions}
\end{align}

\subsection{States}
Symbolic states are conjunctions ($\wedge$) of 
propositions, e.g., $\prop{in box hand} \wedge \prop{above box table} \wedge \neg \prop{on box table}$. PDDL follows the closed-world assumption, which means that propositions that are not explicitly specified are assumed to be false. Thus, we can also represent states as the set of all true propositions, where propositions not in the set are false by default. Let $s$ denote closed-world states, and let $\mathcal{S}$ be the set of all possible states. Note that $|S| = 2^N$.
\begin{align}
    s
        &= \left\{p \in \mathcal{P} \mid p \text{ is true} \right\} \in \mathcal{S}
\end{align}

Under the open-world assumption, propositions that are not explicitly specified in the state are unknown---neither true nor false. Here, symbolic states can be represented as a pair of signed states $(s^+, s^-)$, where $s^+$ is the set of positive propositions, $s^-$ is the set of negative propositions, and propositions that are excluded from $s^+$ and $s^-$ are unknown. Let $\hat{s}$ denote open-world states represented in this manner.
\begin{align}
\begin{split}
    \hat{s}
        &= \left(s^+, s^-\right) \in \mathcal{S}^2 \\
    s^+
        &= \left\{p \in \mathcal{P} \mid p \text{ is true} \right\} \in \mathcal{S} \\
    s^-
        &= \left\{q \in \mathcal{P} \mid \neg q \text{ is true} \right\} \in \mathcal{S}
\end{split}
\end{align}

In practice, we represent closed-world states as boolean vectors $s \in \{0, 1\}^N$ and open-world states as boolean matrices $\hat{s} \in \{0, 1\}^{2 \times N}$, where the first and second row correspond to positive and negative states, respectively.

\subsection{Actions}
Actions are defined by their pre-conditions---a first-order logic formula that must be true before performing the action---and post-conditions---a formula that is applied to the symbolic state after the action. These formulas can include universal (\texttt{forall}), existential (\texttt{exists}), and conditional (\texttt{when}) quantifiers in PDDL. A symbolic planner aims to find a sequence of actions that starts at the initial state $s_{init}$ and ends up in a symbolic state that satisfies the goal $\mathfrak{g}$.

\section{Partial State Labels from Actions}
\label{sec:methods}
\begin{figure*}
    \centering
    \includegraphics[width=\textwidth]{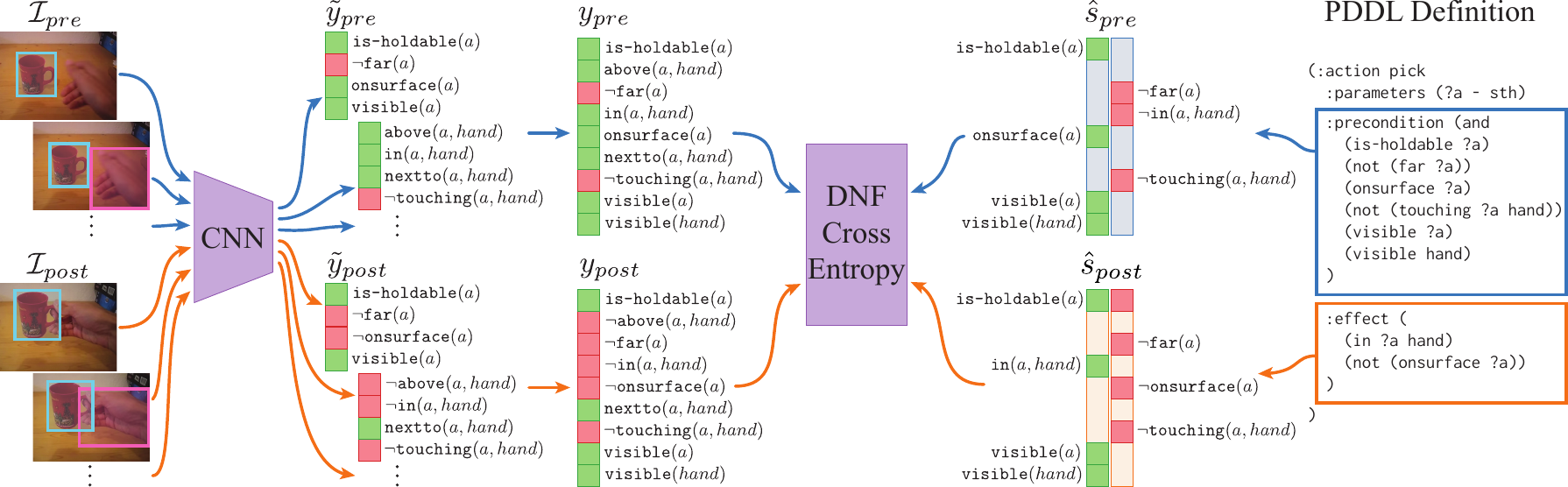}
    \caption{Training example for the 20BN action ``picking [a cup] up". The pre- and post-images $\mathcal{I}_{pre}$ and $\mathcal{I}_{post}$ get fed into the predicate classifier along with all $C$ combinations of $M$ object bounding boxes. Each combination gets mapped to predicates according to the number of arguments (e.g., the cup bounding box gets mapped to $\prop{pred a}$ predicates and the cup and hand together get mapped to $\prop{pred a b}$). The predicate classifier outputs $C$ predictions each for $\tilde{y}_{pre},\, \tilde{y}_{post} \in [0, 1]^P$, where $P$ is the number of predicates in the symbolic domain. These predicate vectors are transformed into symbolic state vectors $y_{pre},\, y_{post} \in [0, 1]^N$, where $N$ is the number of propositions. The ground truth labels used to evaluate these predictions come from partial state labels $\hat{s}_{pre},\, \hat{s}_{post} \in \{0, 1\}^{2 \times N}$, obtained from the PDDL definition of the action $\action{pick a}$ on the right. DNF cross entropy is used to compare the predictions to the partial state labels. At test time, a single image is fed into the network along with bounding boxes for the predicate arguments, and the predicate classifier predicts the symbolic state of the image.}
    \label{fig:2-pipeline}
\end{figure*}

Our goal is to use a PDDL specification to automatically label symbolic states in an entire dataset of videos only annotated with actions. In this section, we formalize our method for obtaining partial state labels from pre- and post-conditions of actions and discuss how to use these partial labels to train symbolic state classifiers. An overview of the pipeline is provided in Fig.~\ref{fig:2-pipeline}.

\subsection{Predicate Classification Network}

Let $M$ be the maximum number of parameters of any predicate in $\mathcal{P}$. Given $\mathcal{I}$ as an RGB image of any dimension, and $M$ bounding boxes $(b_1, \dots, b_M)$ of $M$ objects representing ordered predicate arguments, the network outputs a vector $\tilde{y}$ of $P$ probabilities, one for each predicate in $\mathcal{P}$:
\begin{align}
    \tilde{y} &= f_{network}\left(\mathcal{I}, b_1, \dots, b_M\right) \in [0, 1]^P
\end{align}

The benefit of using object bounding boxes is that we can query the predicate network for propositions with specific argument ordering, such as $\prop{in spoon cup}$ or $\prop{in cup spoon}$. The predicate network would then output a prediction based on its understanding of what it looks like for argument $\obj{a}$ to be $\pred{in}$ argument $\obj{b}$. We assume that most robot manipulation applications will already rely on object tracking for planning and control, and therefore obtaining object bounding boxes for predicate detection would not incur any additional cost. However, using bounding boxes is an implementation choice that is orthogonal to our proposed method of labeling symbolic states with pre- and post-conditions. A classifier that predicts propositions without bounding boxes could also be used in place of the predicate classifier for environments where bounding boxes are difficult to obtain.

\ifx\arxiv\undefined
\else
Details of the network architecture that we use in the experiments can be found in Appx.~\ref{appx:architecture}.
\fi

\subsection{Obtaining Partial State Labels from Actions}
\label{subsec:4b-collapsed-dnf}

The logic formulas that describe pre- and post-conditions can be arbitrarily complex with deeply nested compound formulas. To make these nested formulas usable for computing a neural network loss for symbolic state predictions, we first convert them via first-order logic algebra into a flattened form called disjunctive normal form (DNF). DNFs are written as disjunctions ($\vee$) of conjunctions ($\wedge$) of positive and negative propositions. Each conjunction in a DNF can be interpreted as a partially-specified state under the open-world assumption.
\begin{align}
    DNF
        &= \hat{s}_1 \vee \hat{s}_2 \vee \dots \vee \hat{s}_D
    \label{eq:dnf}
\end{align}

With this interpretation, a pre-condition DNF represents a set of possible partial states before performing the action, and a post-condition DNF represents a set of possible partial states after. These candidate states are partial because actions only specify conditions relevant to the action itself. For example, the action $\action{pick box}$ may require $\neg \prop{in box hand}$ as a pre-condition, but does not care whether $\action{open door}$ is true and is not even aware of $\obj{door}$'s existence. The candidate states in a pre- or post-condition DNF can encode varying amounts of information about the open world (i.e., specifying anywhere from 1 to $N$ propositions), and some candidates may be supersets of others. A DNF is satisfied if and only if at least one of the candidate states in the DNF is true (Eq.~\ref{eq:dnf}). In other words, when an action is performed on a concrete symbolic state, at least one candidate state in the pre-condition DNF will be true before performing the action, and at least one in the post-condition DNF will be true after. Although it is impossible to know which candidate is true without more information, we do know the truth value of the propositions that appear in all the candidate states.

To label symbolic states using pre- and post-conditions, we therefore collapse each DNF into the largest single partial state that satisfies all its conjunctions. This is equivalent to the intersection of all the positive and negative states.
\begin{align}
    \hat{s}_{DNF}
        &= \left(\bigcap_{i=1}^D s^+_i,~ \bigcap_{i=1}^D s^-_i\right) \in \mathcal{S}^2
\end{align}

If collapsing a pre-condition DNF results in the empty set, this means that the pre-conditions are too general to provide labels for symbolic state. Post-conditions with empty collapsed DNFs means that there is a conditional effect (\texttt{when}) that may result in the action producing no changes. Both cases can be avoided by making the pre-conditions more descriptive. For example, if the action $\action{close a}$ contains the conditional post-condition $\prop{is-closable a} \implies \prop{closed a}$, the action can be redefined so that $\prop{is-closable a}$ is a pre-condition and the post-condition is simply $\prop{closed a}$.

Let $\hat{s}_{pre}(a) : \mathcal{A} \rightarrow \mathcal{S}^2$ and $\hat{s}_{post}(a) : \mathcal{A} \rightarrow \mathcal{S}^2$ be functions that return collapsed pre- and post-condition DNFs, respectively, for action $a$. These functions provide a way to label propositions whose values are guaranteed to be known before or after an action is performed. Any other proposition could be either true or false without violating a pre- or post-condition, so nothing can be said about their ground truth values.
\ifx\arxiv\undefined
\else
A formal proof for this statement can be found in Theorem \ref{thm:dnf}.
\fi


\subsection{DNF Cross Entropy Loss}

As summarized in Fig.~\ref{fig:2-pipeline}, a data point in the training set consists of an action $a$ along with its pre- and post-conditions, a pair of images $(\mathcal{I}_{pre}, \mathcal{I}_{post})$ corresponding to the action's before and after state, and bounding boxes of the objects whose predicates one wishes to identify. Let $y_{pre} \in [0, 1]^N$ and $y_{post} \in [0, 1]^N$ be the predicted outputs of the network. The ground truth labels derived from the pre- and post-conditions of $a$ can be represented as a pair of matrices $\hat{s}_{pre}(a),\, \hat{s}_{post}(a) \in \{0, 1\}^{2 \times N}$.

To measure how much a network prediction output agrees with an action's pre- and post-conditions, we define a modified cross entropy loss to handle DNFs. Here, $\sigma(y)$ is the sigmoid function.
\begin{align}
    CE_{DNF}(y, \hat{s})
        &= -s^+ \log \sigma(y) - s^- \log \sigma(-y)
    \label{eq:dnf-ce}
\end{align}

The loss function used to train the network on a pre- and post-condition pair is:
\begin{align}
\begin{split}
    loss(y_{pre}, y_{post}; \hat{s}_{pre}, \hat{s}_{post})
        &= CE_{DNF}(y_{pre}, \hat{s}_{pre}) \\
        &\quad+ CE_{DNF}(y_{post}, \hat{s}_{post})
\end{split}
\end{align}

\section{Experiments}
\label{sec:experiments}
In the following experiments, we apply our framework to a large-scale real-world dataset, where obtaining complete ground truth labels is impractical. We then apply these trained classifiers to perform closed-loop task planning in a real robot environment. Finally, we evaluate the effectiveness of using partial labels of symbolic state from action pre- and post-conditions compared to complete ground truth labels in a toy Gridworld environment.

\subsection{Learning Predicates from Large-Scale Datasets}

The 20BN dataset \cite{goyal2017something} contains $220,847$ video examples of $174$ manipulation actions. Although 20BN does not label bounding boxes for the action arguments, we use the bounding boxes from Something-Else \cite{materzynska2020something}. We have defined the pre- and post-conditions for all but two of the actions---``putting \textit{\#} of \textit{sth} onto \textit{sth}" and ``stacking \textit{\#} of \textit{sth}"---since the \textit{\#} variable cannot be easily described as a PDDL object. We have also defined $35$ predicates relevant to the $172$ actions. The full PDDL description can be found on the \href{https://sites.google.com/stanford.edu/groundingpredicates}{project website}.

\ifx\arxiv\undefined
\ifx\arxiv\undefined
\subsubsection{Back-of-the-envelope calculations}
\label{appx:calculations}

First, we perform back-of-the-envelope calculations to estimate how long it would take to label the symbolic states in 20BN manually compared to our method. 
\else
\subsection{Back-of-the-Envelope Calculations for 20BN}
\label{appx:calculations}

Here, we perform back-of-the-envelope calculations to estimate how long it would take to label the symbolic states in the 20BN dataset manually and using our proposed method.

\fi
Our PDDL specification for 20BN contains $35$ predicates and $151$ total propositions. This means that fully annotating the symbolic state in one image would require choosing the boolean values for $151$ variables. Assuming that an accustomed worker could annotate the entire video in $10$ minutes, labeling all of the $132,853$ videos in our subset of 20BN in would take approximately $22$k working hours, or $920$ days of continuous $24$-hour work. Assuming $8$-hour work days, this would take $2760$ days.

By contrast, suppose defining the pre- and post-conditions of an action takes an expert $10$ minutes. Then, defining the pre- and post-conditions of all $171$ actions in our subset of 20BN would take roughly $30$ hours.

With this method, the entire dataset has been labeled in $4$ $8$-hour work days, or $1/690$ of the time. However, the total number of labeled propositions is also smaller. On average, each action DNF specifies around $37$ propositions with the usage of axioms. This is about $1/4$ of the total propositions, which means the dataset labeled with actions is effectively $4$ times smaller. Manually annotating the full symbolic state of $1/4$ the dataset would still take $690$ $8$-hour work days.
\else
The purpose of this experiment is to show that our framework makes it possible to train symbolic state classifiers on large-scale real world datasets without relying on expensive manual annotations. Back-of-the-envelope calculations in Appx.~\ref{appx:calculations} indicate that using pre- and post-conditions to partially label the symbolic states in all $132,853$ videos of our subset of 20BN would take an expert roughly four $8$-hour work days. By contrast, obtaining the equivalent partial state labels with manual annotation would take $690$ $8$-hour work days.
\fi

\subsubsection{Setup}

We train the predicate classifiers using DNF cross entropy (\textbf{DNF CE}), as well as a weighted version of cross entropy (\textbf{DNF WCE}) using Class-Balanced Loss \cite{cui2019class} to overcoming imbalances in the predicate distribution. There is no oracle because we do not have ground truth labels of the full symbolic state in this real-world dataset.

\subsubsection{Results}


Both \textbf{DNF CE} and \textbf{DNF WCE} are able to learn the predicates specified in the 20BN PDDL with $0.96$ train F1 and $0.92$ test F1. 
Although the two achieve the same test F1 scores (within $0.001$), the predicate F1 scores in Table~\ref{table:20bn-predicate-f1} reveal that overall, \textbf{DNF WCE} performs better on the less frequent predicates, increasing the average predicate F1 score from $0.60$ to $0.65$.

Given that each proposition in 20BN is positive $36\%$ of the time on average, a random classifier that outputs positive propositions with $0.36$ chance would be expected to get $0.36$ F1. Thus, an F1 score of $0.65$ is significantly better than random. Both \textbf{DNF CE} and \textbf{DNF WCE} achieve $0.93$ test accuracy. Fig.~\ref{fig:1-teaser} shows qualitative examples of \textbf{DNF WCE}.

\begin{table}
    \setlength{\belowcaptionskip}{-2mm}
    \centering
    \tiny
\begin{tabular}{| l | c | c c c | c c c |}
\hline
\multirow{2}{*}{Predicate}
                 & \multirow{2}{*}{Dist}
                        & \multicolumn{3}{c|}{\textbf{DNF CE}} & \multicolumn{3}{c|}{\textbf{DNF WCE}} \\
\cline{3-8}
                 &      & Prec. & Rec.  & F1            & Prec. & Rec.  & F1 \\
\hline
is-bendable(a)   & 0.00 & 0.95 & 0.90 & \textbf{0.93} & 0.97 & 0.82 & 0.89          \\
is-fluid(a)      & 0.03 & 0.58 & 0.07 & 0.13          & 0.50 & 0.20 & \textbf{0.28} \\
is-holdable(a)   & 0.06 & 1.00 & 1.00 & 1.00          & 1.00 & 1.00 & \textbf{1.00} \\
is-rigid(a)      & 0.02 & 0.86 & 0.93 & \textbf{0.90} & 0.87 & 0.91 & 0.89          \\
is-tearable(a)   & 0.00 & 0.89 & 0.85 & \textbf{0.87} & 0.88 & 0.84 & 0.86          \\
above(a, b)      & 0.03 & 0.84 & 0.58 & \textbf{0.68} & 0.84 & 0.57 & 0.68          \\
attached(a, b)   & 0.04 & 0.62 & 0.25 & \textbf{0.35} & 0.34 & 0.11 & 0.16          \\
behind(a, b)     & 0.03 & 0.62 & 0.54 & 0.58          & 0.59 & 0.61 & \textbf{0.60} \\
broken(a)        & 0.01 & 0.83 & 0.22 & 0.35          & 0.62 & 0.46 & \textbf{0.53} \\
close(a)         & 0.01 & 0.91 & 0.95 & \textbf{0.93} & 0.92 & 0.90 & 0.91          \\
closed(a)        & 0.00 & 0.65 & 0.76 & 0.70          & 0.65 & 0.79 & \textbf{0.71} \\
deformed(a)      & 0.01 & 0.54 & 0.05 & 0.09          & 0.30 & 0.39 & \textbf{0.34} \\
empty(a)         & 0.00 & 0.62 & 0.44 & 0.52          & 0.45 & 0.72 & \textbf{0.56} \\
far(a)           & 0.08 & 0.17 & 0.03 & 0.05          & 0.17 & 0.15 & \textbf{0.16} \\
fits(a, b)       & 0.00 & 0.98 & 1.00 & \textbf{0.99} & 0.99 & 0.95 & 0.97          \\
folded(a)        & 0.02 & 0.57 & 0.27 & 0.37          & 0.43 & 0.55 & \textbf{0.48} \\
has-hole(a)      & 0.00 & 0.75 & 0.50 & \textbf{0.60} & 0.60 & 0.50 & 0.55          \\
high(a)          & 0.01 & 0.63 & 0.33 & 0.43          & 0.57 & 0.44 & \textbf{0.50} \\
in(a, b)         & 0.10 & 0.90 & 0.88 & \textbf{0.89} & 0.90 & 0.88 & 0.89          \\
infront(a, b)    & 0.03 & 0.65 & 0.51 & 0.57          & 0.63 & 0.54 & \textbf{0.58} \\
left(a)          & 0.00 & 0.60 & 0.68 & 0.64          & 0.60 & 0.70 & \textbf{0.65} \\
low(a)           & 0.01 & 0.65 & 0.62 & \textbf{0.64} & 0.65 & 0.63 & 0.64          \\
nextto(a, b)     & 0.02 & 0.74 & 0.54 & 0.63          & 0.73 & 0.56 & \textbf{0.64} \\
on(a, b)         & 0.04 & 0.80 & 0.52 & \textbf{0.63} & 0.79 & 0.51 & 0.62          \\
onsurface(a)     & 0.02 & 0.87 & 0.93 & \textbf{0.90} & 0.88 & 0.91 & 0.90          \\
open(a)          & 0.00 & 0.63 & 0.50 & 0.56          & 0.63 & 0.54 & \textbf{0.58} \\
right(a)         & 0.00 & 0.63 & 0.53 & 0.58          & 0.64 & 0.53 & \textbf{0.58} \\
stacked(a)       & 0.00 & 0.81 & 0.69 & 0.75          & 0.86 & 0.75 & \textbf{0.80} \\
stretched(a)     & 0.02 & 0.79 & 0.22 & 0.35          & 0.27 & 0.59 & \textbf{0.37} \\
torn(a)          & 0.02 & 0.80 & 0.34 & 0.47          & 0.63 & 0.53 & \textbf{0.57} \\
touching(a, b)   & 0.16 & 0.91 & 0.89 & \textbf{0.90} & 0.90 & 0.88 & 0.89          \\
twisted(a)       & 0.02 & 0.61 & 0.12 & 0.19          & 0.42 & 0.51 & \textbf{0.46} \\
under(a, b)      & 0.03 & 0.83 & 0.58 & 0.69          & 0.82 & 0.60 & \textbf{0.69} \\
upright(a)       & 0.01 & 0.78 & 0.71 & \textbf{0.74} & 0.64 & 0.80 & 0.71          \\
visible(a)       & 0.14 & 1.00 & 1.00 & 1.00          & 1.00 & 1.00 & \textbf{1.00} \\
\hline
\textbf{Average} & 0.03 & 0.75 & 0.54 & 0.60          & 0.66 & 0.61 & \textbf{0.65} \\
\textbf{Overall} & 1.00 & \textbf{0.93} & 0.90 & \textbf{0.92}          & 0.93 & \textbf{0.90} & 0.92 \\
\hline
Random           &      & 0.36 & 0.36 & 0.36          & 0.36 & 0.36 & 0.36 \\
\hline
\end{tabular}
    \caption{20BN precision, recall, and F1 test scores per predicate. The \textit{Dist} column shows each predicate's proportional representation in the predicate distribution. Overall, \textbf{DNF WCE} improves the performance of rare predicates, increasing the average F1 score across all predicates from $0.60$ to $0.65$. This result emphasizes the importance of mitigating skewed predicate distributions common to symbolic domains. The \textit{Overall} row shows the F1 score over the entire test set (not split by predicate). The \textit{Random} row shows the expected performance of a random classifier, given that each proposition is positive $36\%$ of the time on average.}
    \label{table:20bn-predicate-f1}
\end{table}

\subsubsection{Transfer to Real Robot Domains}

\begin{figure}
    \centering
    \includegraphics[width=\columnwidth]{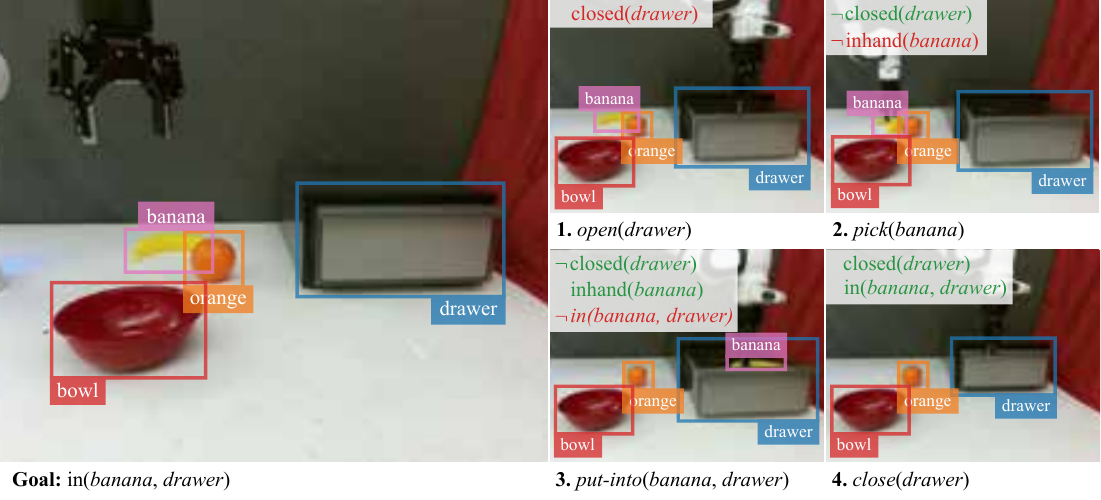}
    \caption{20BN predicate classifier applied to robotic pick-and-place domain. In this closed-loop task planning example, a user first gives the task planner a high-level goal in the form of a logical formula such as $\pred{in}(\obj{banana}, \obj{drawer})$. The predicate classifier then predicts the symbolic state of the environment from RGB image data and sends it to a task planner. The task planner performs symbolic tree search to produce a sequence of actions to accomplish the goal, using 20BN actions such as ``open \textit{sth}" or ``put \textit{sth} into \textit{sth}". As the actions are executed by low-level controllers, the task planner continues to re-plan based on updated symbolic state predictions given by the predicate classifier.}
    \label{fig:3-robot}
\end{figure}

To demonstrate the zero-shot transferability of predicate classifiers trained with our framework to robot manipulation domains, we directly apply the \textbf{DNF WCE} classifier trained on 20BN to a robot pick-and-place setting for closed-loop task planning (Fig.~\ref{fig:3-robot}).

For our real robot experiment, we create a pick-and-place environment where the robot needs to clear fruit off the table and store them in appropriate containers. We use the YOLOv5 object detector \cite{jocher2021yolov5} for predicate bounding boxes along with Mediapipe object tracking \cite{lugaresi2019mediapipe} for smoother tracking between frames. 
The predicate classifier then sends a symbolic state prediction to a task planner that searches for a sequence of actions that satisfies the PDDL goal. The whole pipeline (from perception to task planning) operates at 10Hz and therefore allows closed-loop task planning that can respond to failures and disturbances.
%
A video of this demonstration is provided in the supplementary material.

\subsection{Partial vs. Full State Labels in Gridworld}

Full symbolic state labels are ideal for training predicate classifiers. However, obtaining such labels is only practical for small domains or simulated environments with direct access to the symbolic state. To evaluate the effectiveness of training on partial vs. complete state labels, we use the Gridworld environment (Fig.~\ref{fig:6-gridworld-pred-f1}), where we have full control of the symbolic state.

\subsubsection{Gridworld Environment}


\ifx\arxiv\undefined
In this environment, an agent needs to obtain a trophy from inside a chest. There are $8$ objects and $8$ actions defined with complex pre- and post-conditions, making use of universal (\texttt{forall}), existential (\texttt{exists}), and conditional (\texttt{when}) quantifiers. Details of the environment can be found on the \href{https://sites.google.com/stanford.edu/groundingpredicates}{project website}.
\else
In this environment, an agent needs to obtain a trophy from inside a chest. There are 8 objects: $\obj{agent}$, $\obj{trophy}$, $\obj{chest}$, $\obj{chest\_key}$, $\obj{door}$, $\obj{door\_key}$, $\obj{room\_a}$, and $\obj{room\_b}$. There are 6 predicates:
{\small
\begin{description}[labelindent=2pt,font=\normalfont]
    \item[-- $\prop{reachable a}$:] Whether $\obj{a}$ can be picked up by the agent.
    \item[-- $\prop{closed a}$:] Whether the door/chest $\obj{a}$ is closed.
    \item[-- $\prop{locked a}$:] Whether the door/chest $\obj{a}$ is locked.
    \item[-- $\prop{in a b}$:] Whether $\obj{a}$ is in $\obj{b}$.
    \item[-- $\prop{connects a b c}$:] Whether door $\obj{a}$ connects rooms $\obj{b}$, $\obj{c}$.
    \item[-- $\prop{matches a b}$:] Whether key $\obj{a}$ matches door/chest $\obj{b}$.
\end{description}
}
There are 8 actions to control the $\obj{agent}$:
{\small
\begin{description}[labelindent=2pt,font=\normalfont]
    \item[-- $\action{enter a b}$:] Enter room $\obj{a}$ through door $\obj{b}$.
    \item[-- $\action{goto a b}$:] Go to object $\obj{a}$ in room $\obj{b}$.
    \item[-- $\action{pick a b}$:] Pick up object $\obj{a}$ from object/room $\obj{b}$.
    \item[-- $\action{place a b}$:] Place object $\obj{a}$ inside object/room $\obj{b}$.
    \item[-- $\action{open a}$:] Open door/chest $\obj{a}$.
    \item[-- $\action{close a}$:] Close door/chest $\obj{a}$.
    \item[-- $\action{unlock a b}$:] Unlock door/chest $\obj{a}$ with key $\obj{b}$.
    \item[-- $\action{lock a b}$:] Lock door/chest $\obj{a}$ with key $\obj{b}$.
\end{description}
}

The actions are defined with complex pre- and post-conditions, making use of universal (\texttt{forall}), existential (\texttt{exists}), and conditional (\texttt{when}) quantifiers.
\fi

To generate training data, we first sample a random symbolic state $s_0$, where each proposition in $s_0$ has a $5\%$ chance of being true. There are $172$ propositions in Gridworld, so on average, a randomly sampled state will have $8$ propositions. Then, given an action with its pre- and post-conditions $\hat{s}_{pre}$ and $\hat{s}_{post}$, we generate two symbolic states:
\begin{align}
    s_{pre}
        &= (s_0 \cup s_{pre}^+) - s_{pre}^- \\
    s_{post}
        &= (s_{pre} \cup s_{post}^+) - s_{post}^-
\end{align}

We then render $s_{pre}$ and $s_{post}$ to obtain images $\mathcal{I}_{pre}$, $\mathcal{I}_{post}$.

\subsubsection{Setup}

We compare an \textbf{Oracle} trained on full state labels to two models. The first, \textbf{DNF CE}, is trained with DNF cross entropy (Eq.~\ref{eq:dnf-ce}). Because partial labels contain less information than full labels, we expect \textbf{DNF CE} to take longer to train than \textbf{Oracle} and/or perform slightly worse.

The third model is an ablation study to test whether the model benefits from seeing the \textit{visual change} induced by an action in order to visually ground symbols. This ablation, \textbf{Half DNF}, trains \textbf{DNF CE} on only the pre- or post-conditions of any given action instance, but not both. With normal DNF training, a single data point comes with a pair of images: one before and one after the action. For the ablation, we feed the model with one image from the pair: either before or after. Across all the data, the model will see examples of both the pre- and post-conditions for each action, but never both for a single action \textit{instance}. Because this ablation only sees half the data, we double the size of its dataset for a fair comparison. We expect \textbf{Half DNF} to perform worse than \textbf{DNF CE}.

\textbf{Oracle} and \textbf{DNF CE} are trained on $10,000$ examples of actions, and \textbf{Half DNF} is trained on $20,000$. All of them are evaluated against the full ground truth state (as opposed to the partial DNF state) on the same test set of $10,000$ examples. Using the full state allows us to get a clearer picture of the generalization abilities of these models.

\begin{figure}
    \centering
    \includegraphics[height=.955in]{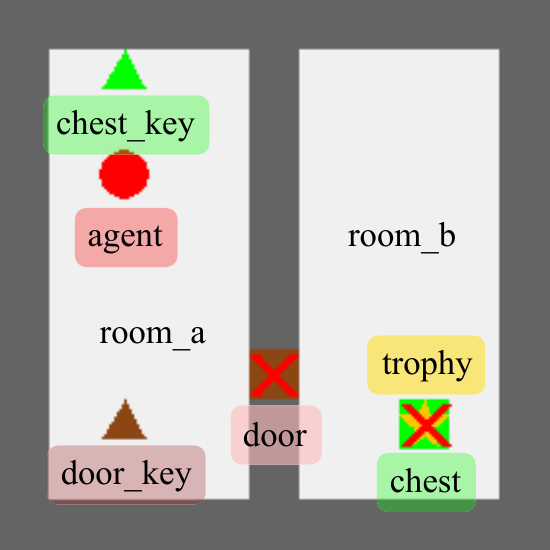}
    \includegraphics[height=.955in]{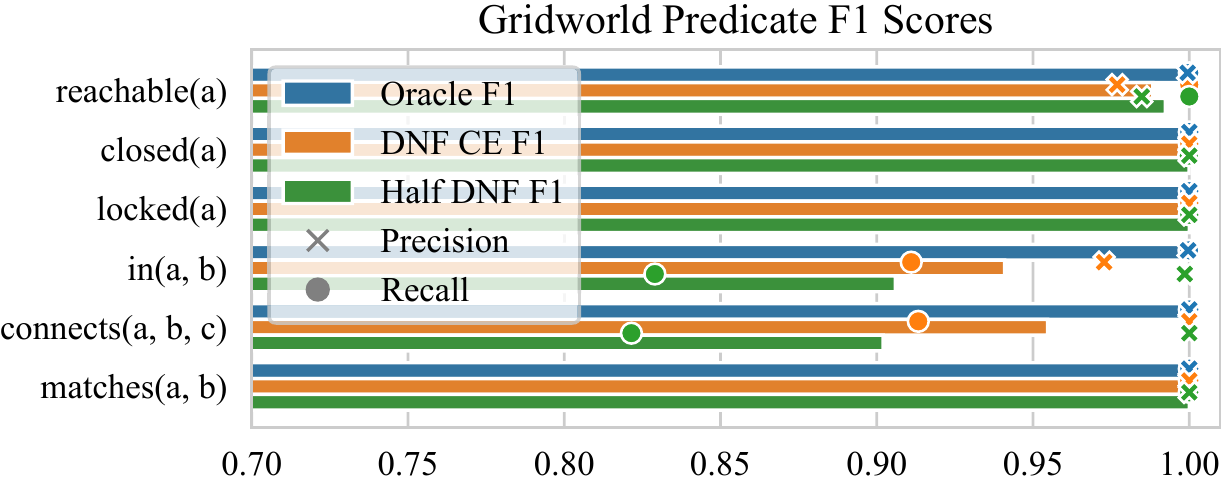}
    \caption{\textit{Left:} In the Gridworld environment, the agent needs to obtain a $\obj{trophy}$ that is locked inside a $\obj{chest}$ locked inside another room.
    \ifx\arxiv\undefined
    \else
    The full PDDL specification can be found on the \href{https://sites.google.com/stanford.edu/groundingpredicates}{project website}.
    \fi
    \textit{Right:} Test F1 scores per predicate for the Gridworld experiment.
    }
    \label{fig:3-gridworld}
    \label{fig:5-gridworld-f1}
    \label{fig:6-gridworld-pred-f1}
\end{figure}

\subsubsection{Results}

As shown in Fig.~\ref{fig:5-gridworld-f1}, all the models achieve a training F1 score of nearly 1, indicating that they finish learning by 20 epochs. \textbf{Oracle} achieves a test F1 of 1.00, \textbf{DNF CE} achieves 0.96, and \textbf{Half DNF} achieves 0.94. \textbf{Half DNF}'s worse performance indicates that seeing the visual changes induced by each action is more beneficial than simply receiving more data (twice the amount).

Increasing the dataset size from $10,000$ to $100,000$ results in perfect test scores for all the DNF models for all the predicates (results not shown). Because DNF training uses partial labels, it requires more data to match the effectiveness of training on complete labels. However, in practice, acquiring a large dataset of actions should be easier than acquiring a small dataset annotated with complete symbolic states (\appx\ref{appx:calculations}). Furthermore, in robotics applications where symbolic state classifiers might be used to determine whether the pre- or post-conditions of an action are satisfied, being able to classify propositions outside of the contexts provided by DNF labels may not be necessary.


\section{Conclusion}
\label{sec:conclusion}
In this work, we have presented a framework for extracting partial symbolic states from action pre- and post-conditions, which can be used to label large datasets with less effort. This new method drastically reduces the cost of labeling large-scale real-world datasets ($690$ vs. $4$ work days for 20BN). Yet, predicate classifiers trained with our method are still able to nearly match the performance of models trained with full ground truth labels, as shown with our Gridworld experiment. Our closed-loop task planning example demonstrates that predicate classifiers trained on large-scale real-world datasets can be applied to real robot domains. In many cases, predicates learned from a general large-scale dataset may not be applicable to custom task planning domains, where accuracy is critical. However, our labeling framework would perhaps be most useful for these very applications where collecting new datasets is necessary. For these situations, our framework makes obtaining real-world symbolic state classifiers much more feasible.

This work opens up many opportunities for long horizon planning in the real world. Other interesting avenues of work include using natural language to learn the pre- and post-conditions of actions \cite{kollar2014grounding,misra2015environment}, or combining visual groundings of symbols with natural language \cite{mao2019neuro,bisk2020experience}.

\section*{Acknowledgment}

Toyota Research Institute (``TRI") provided funds to assist the authors with their research but this article solely reflects the opinions and conclusions of its authors and not TRI or any other Toyota entity.

{\footnotesize
\bibliographystyle{IEEEtranN}
\bibliography{references}
}

\ifx\arxiv\undefined
\else
\appendix

\subsection{Collapsed DNF Proof}
\label{appx:proof}

Here, we provide a proof to justify our proposed formula for collapsing a DNF into a single partial state that can be used to evaluate a symbolic state classifier's predictions (Eq.~\ref{eq:dnf}). Specifically, we require that the formula captures only the propositions whose truth values are guaranteed to be known before or after an action is performed, and that the truth values of propositions outside of the collapsed set cannot be known without additional prior knowledge about the symbolic state.

\begin{theorem}
\label{thm:dnf}
$\hat{s}_{DNF}$ is the largest set of propositions that is fully determined by a DNF.
\end{theorem}

\text{Two prove this statement, we first introduce two lemmas.}

\begin{lemma}
\label{lemma:dnf1}
If a DNF $\hat{s}_1 \vee \hat{s}_2 \vee \dots \vee \hat{s}_D$ is true, then its collapsed form $\hat{s}_{DNF}$ is also true.

\begin{proof}
If the DNF is true, then at least one of $\hat{s}_1, \hat{s}_2, \dots, \hat{s}_D$ must be true. Let $\hat{s}_*$ be one of these true terms. Because $\hat{s}_{DNF}$ is taken as the intersection of $\hat{s}_1, \hat{s}_2, \dots, \hat{s}_D$, it is a subset of $\hat{s}_*$. This means $\hat{s}^*$ can be written as $\hat{s}_{DNF} \wedge \hat{s}_{*-DNF}$, where $\hat{s}_{*-DNF}$ consists of all the positive and negative propositions in $\hat{s}_*$ that are not in $\hat{s}_{DNF}$. Since $\hat{s}_*$ is true, then $\hat{s}_{DNF}$ must also be true.
\end{proof}
\end{lemma}

\begin{lemma}
\label{lemma:dnf2}
If a proposition $p$ is not in the set $\hat{s}_{DNF}$, then both $p$ and $\neg p$ satisfy the DNF $\hat{s}_1 \vee \hat{s}_2 \vee \dots \vee \hat{s}_D$.

\begin{proof}
Suppose for the sake of contradiction that either $p$ or $\neg p$ violates the DNF. Let us consider the case where $p$ violates the DNF. This means that $\neg p$ is required to satisfy the DNF, and that $s^-_1, s^-_2, \dots, s^-_D$ all contain $p$. The intersection of $\hat{s}_1, \hat{s}_2, \dots, \hat{s}_D$ would then contain $p$, and thus $p$ would be in the set $\hat{s}_{DNF}$. However, $p \notin \hat{s}_{DNF}$ by definition, so $p$ must not violate the DNF. The same argument applies for $\neg p$. Neither $p$ nor $\neg p$ can violate the DNF, so both satisfy it.
\end{proof}
\end{lemma}

With these two results, the proof for Theorem \ref{thm:dnf} is straightforward.

\begin{proof}
Lemma \ref{lemma:dnf1} shows that if the DNF $\hat{s}_1 \vee \hat{s}_2 \vee \dots \vee \hat{s}_D$ is true, then $\hat{s}_{DNF}$ is also true. Conversely, the DNF cannot be true if $\hat{s}_{DNF}$ is false. This means that the value of each proposition in $\hat{s}_{DNF}$ is determined by the DNF; in order for the DNF to hold true, their values cannot be arbitrarily true or false. Therefore, $\hat{s}_{DNF}$ is a subset of propositions fully determined by the DNF.

Lemma \ref{lemma:dnf2} shows that if $p \notin \hat{s}_{DNF}$, then both $p$ and $\neg p$ satisfy the DNF $\hat{s}_1 \vee \hat{s}_2 \vee \dots \vee \hat{s}_D$. In other words, any proposition $p \notin \hat{s}_{DNF}$ cannot be determined by the DNF. Therefore, $\hat{s}_{DNF}$ must be the largest set of propositions that is fully determined by the DNF.
\end{proof}

\subsection{Predicate Classifier Architecture}
\label{appx:architecture}

We base the network architecture for our predicate classifier on the bounding-box channel network proposed by \citet{inayoshi2020bounding}, a network that achieves state-of-the-art performance on visual relationship detection tasks. The network takes image features from a backbone image classification network (in our case the fourth layer of ResNet-50 \cite{he2016deep}), and normalizes the spatial size of the image features using RoIAlign \cite{he2017mask} with respect to the region of interest of the image, defined in our case to be the smallest box containing the bounding boxes of all the predicate arguments. This aligned image feature (dimension $7 \times 7 \times 1024$) provides visual context for the predicate arguments. Then, spatial features are created as binary image masks (dimension $7 \times 7 \times 256 M$) generated from the $M$ predicate argument bounding boxes, indicating the locations of each object in the aligned image feature. While \cite{inayoshi2020bounding} populates the spatial features with word vectors corresponding to the name of each object, we simply use a $256$-dimensional vector of ones, since language grounding is outside the scope of this paper.

Together, the image and spatial features allow the network to recognize spatial relationships between arguments. However, in the case where the objects are far apart from each other, their presence in the aligned image feature may be too small. Furthermore, not all $n$-ary predicates depend on spatial relationships (e.g., whether an object can fit inside another). To mitigate these issues, we also provide object image features (dimension $7 \times 7 \times 1024 M$) of each predicate argument normalized with RoIAlign to the argument's bounding box.

The image, spatial, and object features are concatenated along the channel axis (dimension $7 \times 7 \times 1024 + 256 M + 1024 M$) and fed into a convolutional neural network (in our case a ResNet-50 layer) to output a $P$-dimensional vector containing classification probabilities for each of the $P$ predicates.

\fi

\end{document}